\documentclass[letterpaper]{article} 
\usepackage{aaai2026}  
\usepackage{times}  
\usepackage{helvet}  
\usepackage{courier}  
\usepackage[hyphens]{url}  
\usepackage{graphicx} 
\usepackage{natbib}  
\usepackage{caption} 
\usepackage{algorithm}
\usepackage{algorithmic}

\usepackage[style=american]{csquotes}
\MakeOuterQuote{"}

\usepackage{placeins} 
\usepackage{amsmath}
\usepackage{amssymb}
\usepackage{booktabs}
\usepackage{multirow}
\usepackage{subcaption}
\usepackage{tcolorbox}
\tcbuselibrary{skins,breakable}
\usepackage{microtype}
\usepackage{newfloat}
\usepackage{listings}
\DeclareCaptionStyle{ruled}{labelfont=normalfont,labelsep=colon,strut=off} 
\floatstyle{ruled}
\newfloat{listing}{tb}{lst}{}
\floatname{listing}{Listing}
\definecolor{boxBG}{HTML}{ffffff}      
\definecolor{modelRed}{HTML}{C51D1D}   
\definecolor{outerBG}{HTML}{FFFFFF}   
\definecolor{stripBG}{HTML}{C4C4C4}   
\definecolor{figureBoxFrame}{HTML}{1b6102} 
\definecolor{modelBoxFrame}{HTML}{ED9121} 
\definecolor{modelBlue}{HTML}{0F5CC0}
\definecolor{modelGreen}{HTML}{157F3B}
\definecolor{modelOrange}{HTML}{C65A1E}
\definecolor{shade}{RGB}{235,245,235}

\tcbset{
  figurebox/.style={
    enhanced,
    width=\linewidth,
    colframe=figureBoxFrame,
    colback=boxBG,      
    boxrule=1pt,
    arc=6pt,
    left=6pt,right=6pt,top=3pt,bottom=1pt,
  },
    outerbox/.style={
    enhanced,
    width=\linewidth,
    colback=outerBG,
    colframe=black,
    arc=12pt,
    boxrule=0.8pt,
    left=6pt,right=6pt,top=6pt,bottom=6pt,
    drop shadow={
      shadow scale=1,
      shadow xshift=1.2ex,
      shadow yshift=-1.2ex,
      opacity=0.15},
  },
videostrip/.style={
  enhanced,
  colback=stripBG,
  colframe=stripBG,
  boxrule=0pt,
  arc=0pt,
  left=0pt,right=0pt,top=0pt,bottom=0pt,
  overlay={
    \fill[white]
      (frame.north west) rectangle ++ (\linewidth,-3pt);
    \fill[white]
      (frame.south west) rectangle ++ (\linewidth, 3pt);
  },
},
  modelbox/.style={
    enhanced,
    colback=white,
    colframe=modelBoxFrame,
    boxrule=1pt,
    arc=6pt,
    left=6pt,right=6pt,top=3pt,bottom=3pt,
  }
}

\nocopyright

\title{ReasonAct: Progressive Training for Fine-Grained Video Reasoning \\in Small Models}

\author{
    Jiaxin Liu\textsuperscript{\rm 1},
    Zhaolu Kang\textsuperscript{\rm 2}
}
\affiliations{
    \textsuperscript{\rm 1}Siebel School of Computing and Data Science, University of Illinois Urbana-Champaign\\
    \textsuperscript{\rm 2}School of Software \& Microelectronics, Peking University\\
    jiaxin26@illinois.edu, zlkang25@stu.pku.edu.cn
}

\usepackage{bibentry}

\begin{document}

\maketitle

\begin{abstract}
While recent multimodal models have shown progress in vision-language tasks, small-scale variants still struggle with the fine-grained temporal reasoning required for video understanding. We introduce \textbf{ReasonAct}, a method that enhances video reasoning in smaller models through a three-stage training process: first building a foundation with text-only reasoning, then fine-tuning on video, and finally refining with temporal-aware reinforcement learning. We build upon Temporal Group Relative Policy Optimization (T-GRPO) by incorporating temporal consistency modeling into policy optimization. We also propose a \textbf{biomechanically-motivated sub-action decomposition mechanism} that provides graduated rewards for constituent action phases. Through experiments on HMDB51, UCF-101, and Kinetics-400, our 3B-parameter model achieves 67.2\%, 94.1\%, and 78.9\% accuracy respectively, demonstrating improvements of 17.9, 15.8, and 12.3 points over baselines. Ablation studies validate that our progressive training enables smaller models to achieve competitive video reasoning performance while maintaining computational efficiency. 
\end{abstract}


\section{Introduction}
Large multimodal language models have shown strong performance on vision-language tasks~\citep{achiam2023gpt, team2023gemini, liu2024visual}. However, deploying these models in resource-constrained environments is challenging due to high computational requirements. Small-scale multimodal models (1-7B parameters) face fundamental parameter allocation challenges, requiring efficient distribution across visual encoding, language understanding, temporal modeling, and reasoning capabilities, particularly limiting their performance in complex temporal reasoning tasks.

Video understanding is particularly challenging for multimodal reasoning, as it requires models to integrate spatial visual comprehension with temporal dynamics, causal reasoning, and sequential pattern recognition. Traditional approaches to video understanding have predominantly focused on feature extraction and pattern matching, treating actions as atomic classification units without explicit modeling of their internal temporal structure~\citep{feichtenhofer2019slowfast, lin2019tsm, bertasius2021space}. This paradigm fails to capture the fine-grained, hierarchical nature of human actions, where complex behaviors emerge from sequences of coordinated sub-actions with specific temporal dependencies~\citep{winter2009biomechanics,bartlett2007introduction,schmidt2011motor}.

Consider the seemingly simple task of recognizing a "jump" action in a video. While existing models might successfully classify this action based on learned visual patterns, they lack understanding of the underlying biomechanical sequence—how preparation, loading, propulsion, and flight phases interconnect and depend on each other. This understanding helps with recognition and reasoning about why actions happen and how they connect over time.

In contrast to large-scale models with ample capacity, small models must strategically allocate limited resources across modalities. Moreover, existing pre-training paradigms for multimodal models often prioritize vision-language alignment over reasoning development, resulting in models that excel at describing visual content but struggle with complex inferential processes.

Current approaches to enhancing reasoning in multimodal models typically rely on either scaling model parameters~\citep{chowdhery2022palm, hoffmann2022training} or leveraging extensive supervision from superior models~\citep{taori2023stanford, chiang2023vicuna}. However, these strategies are computationally expensive and may not address the fundamental architectural and training methodological challenges faced by smaller models. Self-improvement techniques such as Self-Taught Reasoner~\citep{zelikman2022star} and iterative refinement methods~\citep{huang2022large} offer promising alternatives, but our preliminary experiments indicate that these approaches provide limited benefits when applied directly to temporal video reasoning tasks in resource-constrained settings.

These observations motivate us to explore an alternative approach that systematically builds reasoning capabilities without requiring massive scale. To overcome these limitations in smaller models, we developed \textbf{ReasonAct}. Our approach develops reasoning skills through three stages: foundational training, video fine-tuning, and reinforcement learning with sub-action guidance. Our progressive multi-stage training paradigm builds reasoning capabilities incrementally through Foundational Reasoning Enhancement (FRE) that establishes basic reasoning patterns using diverse text-only tasks, Video-Specific Chain-of-Thought Fine-tuning that adapts these reasoning capabilities to temporal visual content, and Temporal-Aware Reinforcement Learning that refines reasoning strategies through structured reward optimization. 

We adapt Temporal Group Relative Policy Optimization (T-GRPO)~\citep{feng2025videor1} by incorporating new temporal consistency components, creating a temporal-aware extension suitable for fine-grained video reasoning. We propose a biomechanically-motivated sub-action recognition framework that breaks complex actions into constituent sub-actions based on biomechanical analysis, providing graduated rewards during training that enable models to develop fine-grained understanding of action mechanics and temporal dependencies.

Our key contribution is a progressive training methodology that enables 3B-parameter models to achieve competitive video reasoning performance. Through the synergy of foundational reasoning enhancement, video-specific fine-tuning, and our sub-action decomposition within an enhanced T-GRPO framework, ReasonAct achieves 67.2\% on HMDB51~\citep{kuehne2011hmdb}, 94.1\% on UCF-101~\citep{soomro2012ucf101}, and 78.9\% on Kinetics-400~\citep{kay2017kinetics}—improvements of 17.9, 15.8, and 12.3 points respectively over strong baselines. These results demonstrate that this training methodology can largely close the gap between small and large models in video understanding, offering a practical path for resource-constrained deployment.

\begin{figure*}[t]\centering
\includegraphics[width=\linewidth]{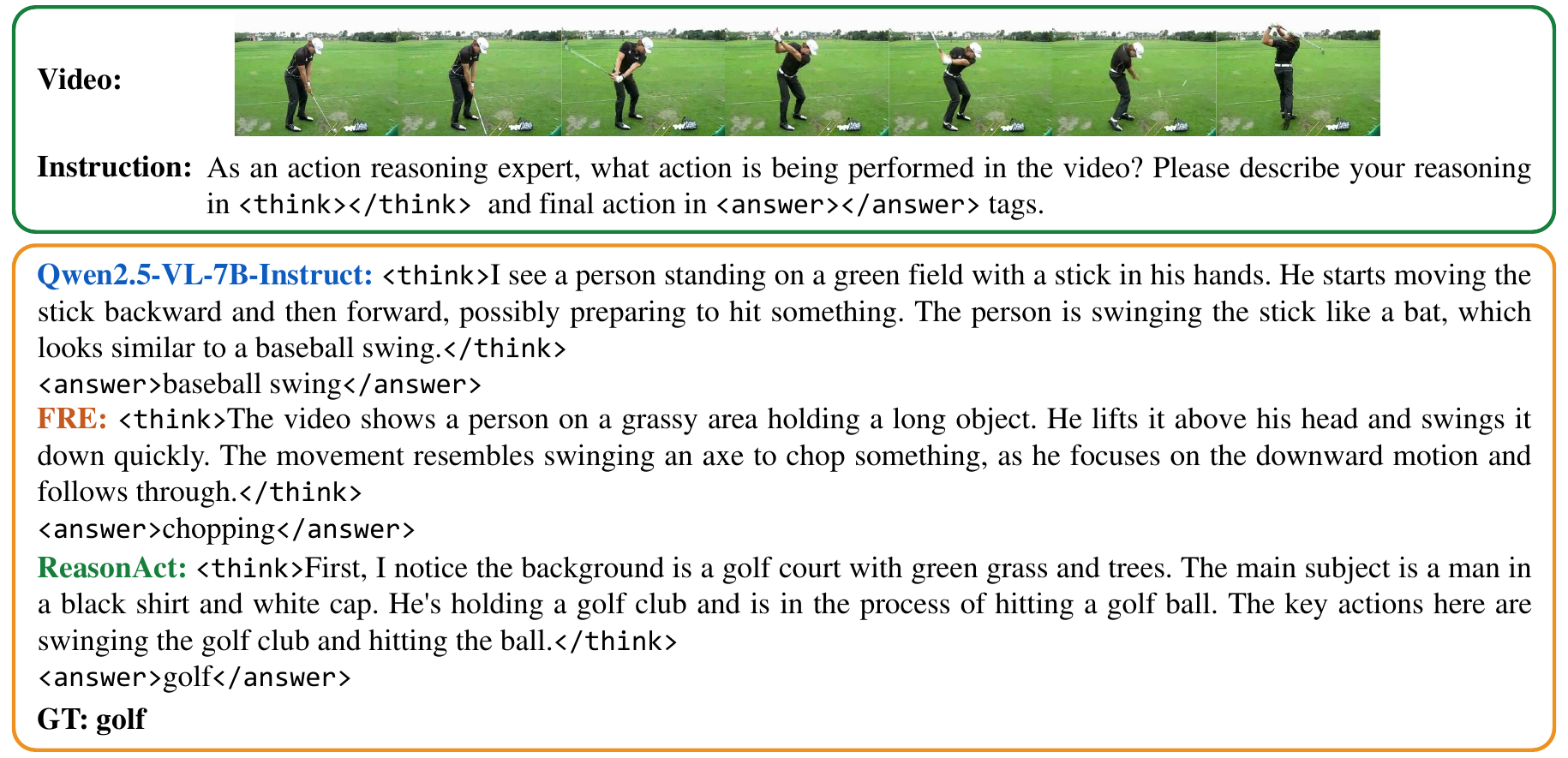}
\captionsetup{justification=centering}
\caption{Example of action reasoning. Baseline Qwen2.5-VL misidentifies golf as baseball swing, FRE incorrectly predicts chopping, while ReasonAct correctly identifies golf through structured reasoning.}
\end{figure*}

\section{Related Work}

\subsection{Video Understanding and Temporal Reasoning}

Video understanding has evolved from traditional approaches using hand-crafted features~\citep{wang2011action, laptev2005space} to modern deep learning architectures~\citep{karpathy2014large}. Recent transformer-based models have shown promising results~\citep{bertasius2021space, liu2022video, yan2022multiview, srivastava2024omnivec2}, yet most approaches treat actions as atomic classification units without explicit temporal structure modeling. This limitation becomes particularly apparent when dealing with complex actions that require understanding of internal phases and temporal dependencies.

Temporal reasoning in video understanding requires capturing long-range dependencies and causal relationships~\citep{zhou2018temporal,fateh2025videollm}. While some works explore hierarchical action recognition~\citep{zhao2017temporal, tang2019coin}, they primarily focus on long-duration activities rather than fine-grained sub-action decomposition for individual action instances. Our approach differs by focusing on sub-action decomposition that enhances reasoning capabilities rather than simply improving classification accuracy. 

\subsection{Multimodal Language Models and Reasoning}

The development of multimodal language models has progressed from early vision-language fusion approaches~\citep{antol2015vqa, anderson2018bottom} to sophisticated architectures capable of complex reasoning~\citep{li2022blip, li2023blip, dai2023instructblip}. However, most existing models prioritize vision-language alignment over reasoning development, particularly in smaller-scale variants where parameter budget constraints become critical~\citep{li2025miv}.

Recent work has identified significant reasoning gaps between multimodal models and their language-only counterparts~\citep{zhang2023multimodal}. While some approaches attempt to address this through specialized training~\citep{liu2024visual, zhu2023minigpt, li2025taco}, systematic methodologies for reasoning enhancement in resource-constrained models remain underexplored. Our work addresses this gap by providing a framework specifically designed for small-scale models.

\subsection{Reinforcement Learning for Language Models}

Reinforcement learning has emerged as a powerful paradigm for aligning language model behavior with desired objectives~\citep{ouyang2022training, bai2022training}. Recent advances in policy optimization algorithms, particularly Group Relative Policy Optimization (GRPO)~\citep{shao2024grpo}, have shown effectiveness in enhancing reasoning capabilities. An important extension is T-GRPO, which adapts GRPO for temporal sequences. However, its mechanism primarily encourages general temporal awareness rather than the fine-grained reasoning our work targets. These approaches typically use reward functions based on task accuracy or human feedback to guide model behavior.

However, existing RL approaches for language models primarily focus on text-only domains and lack mechanisms for handling temporal consistency in multimodal sequences. The adaptation of these techniques to video understanding domains presents unique challenges in reward design and temporal modeling. Our adaptation of T-GRPO addresses these challenges by incorporating explicit temporal awareness into policy optimization while maintaining the computational efficiency advantages of GRPO.

\subsection{Action Decomposition and Sub-Action Recognition}

Action decomposition has been explored in computer vision literature, primarily for activity recognition in long videos~\citep{pirsiavash2014assessing, tang2019coin}. However, these approaches differ fundamentally from our work in several key aspects. Traditional methods focus on temporal localization rather than reasoning enhancement, target long-duration activities rather than individual action instances, and use sub-action recognition as an end task rather than a training signal for reasoning development.

Our sub-action decomposition framework integrates insights from biological motion recognition research~\citep{giese2003neural} while integrating this knowledge into a reinforcement learning framework for multimodal reasoning enhancement. This integration enables models to develop fine-grained understanding of action mechanics that supports robust reasoning about complex temporal sequences.

\section{Methodology}

\subsection{Problem Formulation}

We formulate video reasoning enhancement as a sequential decision-making problem where a small-scale multimodal language model must progressively develop capabilities across three dimensions: foundational reasoning, video-specific understanding, and temporal consistency maintenance. Given a video sequence $\mathbf{V} = \{f_1, f_2, \ldots, f_T\} \in \mathbb{R}^{T \times H \times W \times C}$ and a natural language query $q$, the model must generate a structured reasoning trace $\mathbf{r} = \{r_1, r_2, \ldots, r_L\}$ that leads to a correct answer $a$. A high-quality reasoning trace should not only analyze visual content across time but also identify critical sub-actions and coherently connect them to support a sound final inference. 

Our objective is to optimize the model parameters $\theta$ to maximize both task performance and reasoning quality:

$$\mathcal{J}(\theta) = \mathbb{E}_{(\mathbf{V}, q, a^*) \sim \mathcal{D}} \left[ R_{\text{task}}(\mathbf{r}, a, a^*) + \lambda R_{\text{reasoning}}(\mathbf{r}, \mathbf{V}) \right]$$

where $R_{\text{task}}$ measures task accuracy, $R_{\text{reasoning}}$ evaluates reasoning quality including temporal consistency and sub-action recognition, and $\lambda$ balances these objectives.

\begin{figure*}[t]\centering
\includegraphics[width=\linewidth]{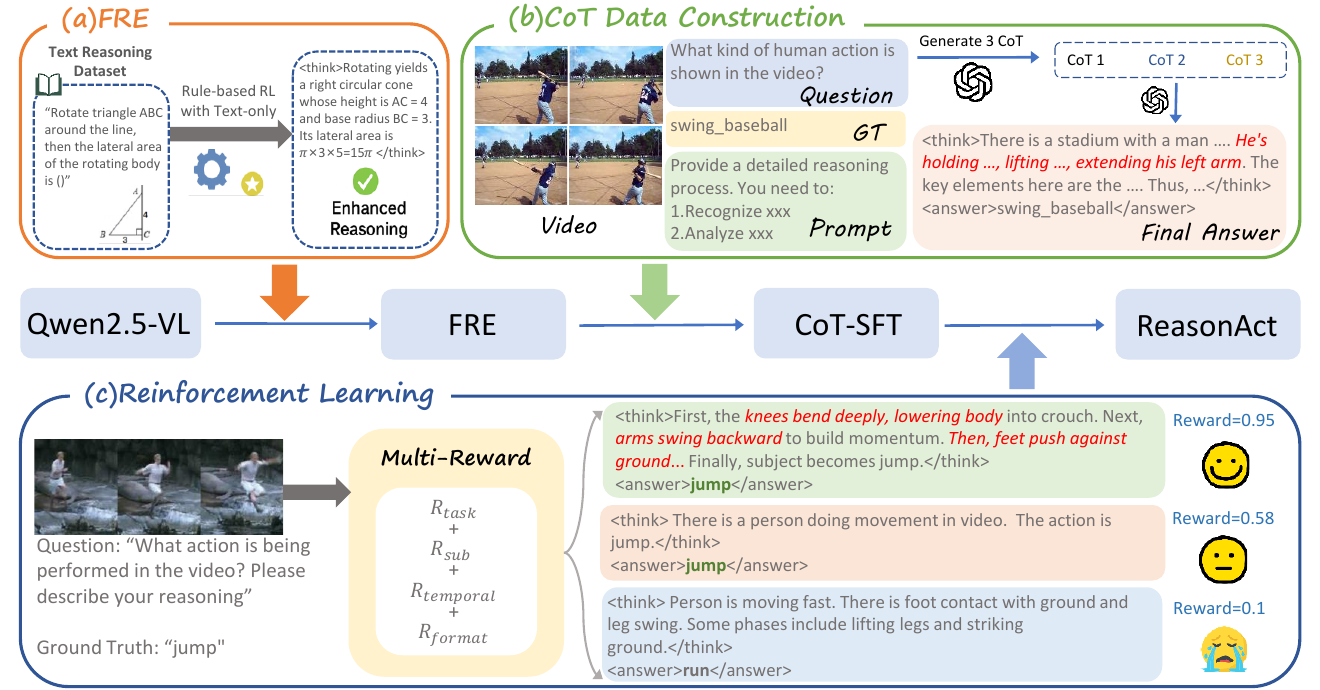}
\captionsetup{justification=centering}
\captionof{figure}{\small Overview of the ReasonAct framework showing the three-stage training paradigm and key technical components.}
\label{fig:framework_overview}
\end{figure*}

To optimize this objective effectively, we must first understand the unique challenges faced by small-scale models. Small models face a key challenge: they must split limited parameters between visual encoding, language understanding, and reasoning. Direct end-to-end training often fails because the model doesn't develop strong reasoning skills. To address this, our ReasonAct framework, as illustrated in Figure~\ref{fig:framework_overview}, implements a three-stage progressive training paradigm that builds capabilities in a curriculum-based manner. Each stage targets specific aspects of the video reasoning task while building upon the foundations established in previous stages, enabling small models to achieve performance comparable to their larger counterparts.

\subsection{Stage 1: Foundational Reasoning Enhancement (FRE)}

Unlike large-scale models that can dedicate substantial parameters to each component, smaller models require careful optimization of this allocation. Our FRE stage addresses this challenge by establishing foundational text-based reasoning skills before introducing the complexity of temporal visual understanding. We fine-tune the base model on a diverse collection of text-only reasoning tasks designed to build foundational cognitive capabilities:

\textbf{Mathematical Reasoning:} Multi-step problem solving requiring logical deduction and calculation chains (e.g., GSM8K~\citep{cobbe2021training}, MATH~\citep{hendrycks2021measuring}).

\textbf{Logical Inference:} Deductive and inductive reasoning tasks that develop systematic thinking patterns (e.g., LogiQA~\citep{liu2020logiqa}, ReClor~\citep{yu2020reclor}).

\textbf{Common-Sense Reasoning:} World knowledge application and causal reasoning tasks (e.g., CommonsenseQA~\citep{talmor2019commonsenseqa}, StrategyQA~\citep{geva2021did}).

\textbf{Structured Analysis:} Tasks requiring systematic decomposition and step-by-step reasoning (e.g., ARC~\citep{clark2018think}, OpenBookQA~\citep{mihaylov2018can}).

The training objective for FRE optimizes cross-entropy loss over reasoning chains:

$$\mathcal{L}_{\text{FRE}} = -\sum_{i=1}^{N} \sum_{j=1}^{L_i} \log P(r_{i,j} | r_{i,<j}, q_i; \theta)$$

where $r_{i,j}$ represents the $j$-th token in the reasoning chain for query $q_i$. With these reasoning foundations established, the model is now prepared to extend its capabilities to temporal visual content, which we address in the next stage.

\subsection{Stage 2: Video-Specific Chain-of-Thought Fine-tuning}

Building upon the reasoning foundations established in Stage 1, we develop video-specific understanding capabilities through supervised fine-tuning with chain-of-thought annotations. This stage teaches the model to apply its reasoning skills to temporal visual content while maintaining the structured thinking patterns learned previously.

We generate chain-of-thought annotations through a structured prompting strategy applied to larger teacher models (GPT-4o). The annotations guide models to systematically analyze visual content across temporal sequences while identifying key visual cues and their temporal relationships. Through these examples, models learn to recognize biomechanical patterns and sub-action sequences. They are also trained to connect temporal observations to a final classification using a structured, logical reasoning process.

The training objective extends the FRE formulation to include video conditioning:

$$\mathcal{L}_{\text{V-SFT}} = -\sum_{i=1}^{N} \sum_{j=1}^{L_i} \log P(r_{i,j} | r_{i,<j}, \mathbf{V}_i, q_i; \theta)$$

Quality control is enforced through multi-stage filtering: automated consistency checking, human expert validation for temporal reasoning accuracy, and iterative refinement based on model feedback during training.

\subsection{Stage 3: Temporal-Aware Reinforcement Learning}
The final stage applies reinforcement learning to refine reasoning quality through four interconnected components: sub-action decomposition, temporal consistency modeling, policy optimization, and integrated reward design.

\subsubsection {Biomechanically-Motivated Sub-Action Recognition}

Traditional video understanding approaches treat actions as atomic units, failing to capture the hierarchical nature of human movement. Our sub-action decomposition framework addresses this limitation by breaking complex actions into constituent phases based on established biomechanical and motor control principles~\citep{winter2009biomechanics,schmidt2011motor}.

\paragraph{Sub-Action Library Construction.}
Based on these principles, we construct sub-action libraries for major action categories. Each action $a$ is decomposed into an ordered sequence of sub-actions $\mathbf{S}_a = \{s_1, s_2, \ldots, s_k\}$ representing key movement phases.

For example, the "jump" action is decomposed as follows:

\begin{itemize}
\item \textbf{Preparation Phase ($s_1$):} "knee flexion", "weight shift", "arm positioning", "stance adjustment"
\item \textbf{Loading Phase ($s_2$):} "deep crouch", "muscle loading", "energy storage", "countermovement"  
\item \textbf{Propulsion Phase ($s_3$):} "explosive extension", "ground contact", "force generation", "takeoff initiation"
\item \textbf{Flight Phase ($s_4$):} "airborne", "body alignment", "trajectory", "landing preparation"
\end{itemize}

This decomposition aligns with established motor control principles that identify distinct movement phases in skilled actions~\citep{winter2009biomechanics}.

To accommodate the diverse ways in which sub-actions may be described, we associate each with multiple semantic variants drawn from real-world reasoning expressions. Our complete library covers 51 action categories from HMDB51, 101 categories from UCF-101, and 400 categories from Kinetics-400, totaling 1,847 unique sub-actions.

\paragraph{Semantic Similarity-Based Detection.}
To detect sub-actions in model-generated reasoning text, we employ a semantic similarity framework using pre-trained sentence embeddings. This approach handles linguistic variability while maintaining precision in sub-action recognition.

For each sub-action $s_i$ with description set $\mathcal{D}_{s_i} = \{d_1, d_2, \ldots, d_m\}$, we compute similarity with reasoning text segment $t$ as:

$$\text{sim}(t, s_i) = \max_{d \in \mathcal{D}_{s_i}} \frac{\mathbf{e}(t) \cdot \mathbf{e}(d)}{|\mathbf{e}(t)| |\mathbf{e}(d)|}$$

where $\mathbf{e}(\cdot)$ represents sentence embeddings from a fine-tuned SentenceTransformer model. Sub-action detection uses an adaptive threshold $\tau_i$ learned during validation to optimize precision-recall trade-offs for each sub-action type.

\paragraph{Graduated Reward Structure.}
The sub-action recognition reward provides graduated feedback based on partial understanding, encouraging models to develop fine-grained action comprehension:

$$\begin{aligned}
R_{\mathrm{sub}}(\mathbf{r},a^*) 
&= \alpha \frac{|\mathcal{S}_{\mathrm{detected}}\cap\mathcal{S}_{a^*}|}
             {|\mathcal{S}_{a^*}|}
  - \beta \frac{|\mathcal{S}_{\mathrm{detected}}\setminus\mathcal{S}_{a^*}|}
               {|\mathcal{S}_{a^*}|} \\
&\quad + \gamma\,P\bigl(\mathcal{S}_{\mathrm{detected}},\mathcal{S}_{a^*}\bigr)
\end{aligned}$$

where $\mathcal{S}_{\text{detected}}$ represents detected sub-actions in the reasoning text, $\mathcal{S}_{a^*}$ represents ground-truth sub-actions for action $a^*$, $P(\cdot, \cdot)$ measures temporal ordering accuracy, and $\alpha, \beta, \gamma$ are reward coefficients optimized through grid search.

\begin{figure}[t]
\centering
\includegraphics[width=\linewidth]{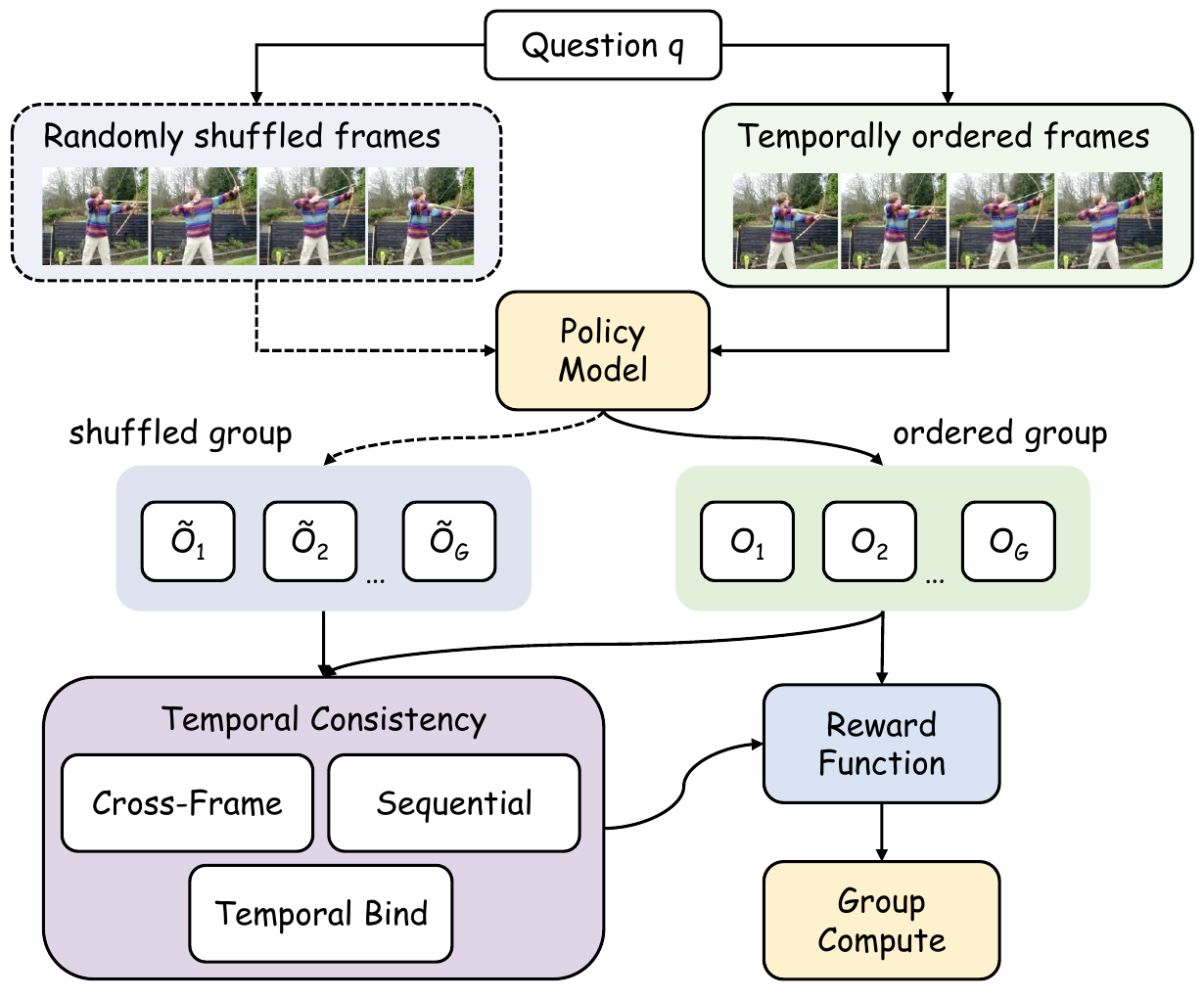}
\caption{Enhanced T-GRPO algorithm flowchart showing temporal consistency modeling and multi-reward integration, compared with standard GRPO.}
\label{fig:tgrpo_flow}
\end{figure}

\begin{table*}[t]
\centering
\caption{Performance comparison across video understanding benchmarks (mean ± std over 5 runs).}
\label{tab:main_results}
\begin{tabular}{l|ccc|ccc}
\toprule
\multirow{2}{*}{Method} & \multicolumn{3}{c|}{Accuracy (\%)} & \multicolumn{3}{c}{Improvement} \\
& HMDB51 & UCF-101 & Kinetics-400 & HMDB51 & UCF-101 & Kinetics-400 \\
\midrule
Qwen2.5-VL-3B (Baseline) & 49.3±1.2 & 78.3±0.8 & 66.6±1.4 & - & - & - \\
\quad + Stage 1 (FRE) & 53.1±1.0 & 82.5±0.7 & 71.2±1.1 & +3.8 & +4.2 & +4.6 \\
\quad + Stage 2 (V-SFT) & 59.4±0.9 & 87.1±0.6 & 75.8±1.0 & +10.1 & +8.8 & +9.2 \\
\quad + Stage 3 (T-GRPO) & 64.7±1.3 & 91.3±1.5 & 77.4±0.9 & +15.4 & +13.0 & +10.8 \\
ReasonAct (Full Model) & \textbf{67.2±0.7} & \textbf{94.1±1.2} & \textbf{78.9±0.8} & \textbf{+17.9} & \textbf{+15.8} & \textbf{+12.3} \\
\midrule
\multicolumn{7}{l}{\textit{Comparison with Recent Video Understanding Models:}} \\
Video-ChatGPT & 54.1±1.4 & 81.2±1.0 & 69.8±1.6 & +4.8 & +2.9 & +3.2 \\
Video-LLaMA & 50.8±1.3 & 79.5±2.1 & 68.4±1.5 & +1.5 & +1.2 & +1.8 \\
LLaVA-Video & 56.6±1.1 & 84.3±0.9 & 72.1±1.2 & +7.3 & +6.0 & +5.5 \\
\bottomrule
\end{tabular}
\end{table*}

\subsubsection{Temporal Consistency Modeling}

While existing T-GRPO approaches provide a foundation for temporal awareness through ordered vs. shuffled frame comparison, this approach primarily ensures general reliance on temporality without fine-grained temporal understanding. To address this limitation, we develop a temporal consistency framework that operates on three interconnected levels, providing denser feedback for precise temporal reasoning.

Sequential coherence ensures that reasoning steps follow logical temporal order and maintain causal relationships between observations. Cross-frame consistency maintains consistent object and action identification across video frames, preventing contradictory interpretations of the same visual elements. Finally, temporal binding correctly associates observed sub-actions with appropriate temporal windows, enabling precise localization of action phases within the video sequence.

To quantify the model's adherence to these principles, we measure temporal consistency through the following score, which averages the scores of the three components:
$$S_{\text{temporal}}(\mathbf{r}, \mathbf{V}) = \frac{1}{3}\left( S_{\text{seq}}(\mathbf{r}) + S_{\text{cross}}(\mathbf{r}, \mathbf{V}) + S_{\text{bind}}(\mathbf{r}, \mathbf{V}) \right)$$

We implement the three temporal consistency components using lightweight Transformer-based classifiers trained on 30k video clips curated from our training datasets, with temporal consistency labels generated using GPT-4o following systematic prompting guidelines. $S_{\text{seq}}$ measures step ordering via Kendall's $\tau$ between predicted and ground-truth temporal sequences, normalized to $[0,1]$. $S_{\text{cross}}$ computes entity consistency using object tracking across frames, while $S_{\text{bind}}$ aligns sub-action spans with temporal windows using IoU averaging. We use clip $\epsilon = 0.2$ and KL coefficient 0.05 for training stability.



\subsubsection{Enhanced Policy Optimization}

Building upon existing T-GRPO approaches, we enhance the policy optimization process by integrating our multi-level temporal consistency modeling with sub-action rewards. Unlike standard T-GRPO which relies primarily on coarse temporal signals, our approach unifies all guiding signals—task accuracy, sub-action recognition, and temporal consistency—into a single, comprehensive reward function to better guide the model.

Our enhanced T-GRPO algorithm refines the model's policy using an objective function based on Proximal Policy Optimization (PPO). This reward is used to compute a temporal-aware advantage function, which quantifies the benefit of taking a specific action at each step. The policy is optimized by maximizing the following objective:

\begin{align*}
\mathcal{L}_{\mathrm{T\text{-}GRPO}}
  &= \mathbb{E}_{t}\Bigl[
       \min\bigl(
         r_t(\theta)\,A_t, \\
  &\qquad\qquad
         \mathrm{clip}\bigl(r_t(\theta),\,1-\epsilon,\,1+\epsilon\bigr)\,A_t
       \bigr)
     \Bigr]
\end{align*}

where $r_t(\theta) = \frac{\pi_{\theta}(a_t|s_t)}{\pi_{\theta_{\text{old}}}(a_t|s_t)}$ is the importance sampling ratio and $A_t$ is the temporal-aware advantage function computed from our integrated reward (detailed in Section 3.3.4). The advantage incorporates temporal consistency through the episode-level reward bonus $S_{\text{temporal}}$, ensuring that reasoning chains with higher temporal coherence receive stronger positive reinforcement during policy updates. This enhancement enables the model to develop fine-grained temporal understanding beyond the general temporality awareness provided by the original T-GRPO.




\subsubsection{Complete Reward Function}

The final reward function integrates multiple components to provide comprehensive feedback on reasoning quality:

$$\begin{aligned}
R_{\mathrm{total}}(\mathbf{r},\mathbf{V},a^*)
&= R_{\mathrm{task}}(a,a^*) + \lambda_1\,R_{\mathrm{sub}}(\mathbf{r},a^*) \\
&\quad + \lambda_2\,S_{\mathrm{temporal}}(\mathbf{r},\mathbf{V})
      + \lambda_3\,R_{\mathrm{format}}(\mathbf{r})
\end{aligned}$$

where $R_{\text{task}}$ measures task accuracy, $R_{\text{sub}}$ evaluates sub-action recognition quality, and $R_{\text{format}}$ ensures proper output formatting. The temporal consistency score $S_{\text{temporal}}(\mathbf{r},\mathbf{V})$ is integrated directly as an episode-level reward component to guide the policy toward generating temporally coherent reasoning. The coefficients $\{\lambda_i\}$ balance these objectives and are optimized through extensive hyperparameter search.

\section{Experiments}

\subsection{Experimental Setup}

\subsubsection{Datasets and Evaluation Metrics}

We evaluate ReasonAct on three established video understanding benchmarks:

\textbf{HMDB51}~\citep{kuehne2011hmdb}: 6,849 clips across 51 action categories, focusing on human motion analysis with complex temporal dynamics.

\textbf{UCF-101}~\citep{soomro2012ucf101}: 13,320 clips spanning 101 diverse action categories, providing full evaluation across different action types.

\textbf{Kinetics-400}~\citep{kay2017kinetics}: 400 action categories with emphasis on temporal reasoning and fine-grained action discrimination.

Evaluation metrics include classification accuracy, reasoning quality scores (combining automated metrics and human evaluation), temporal consistency measures, and computational efficiency analysis.

\subsubsection{Implementation Details}

We use Qwen2.5-VL-3B-Instruct as our base model, sampling 16 uniformly distributed frames during both training and inference. The model architecture remains unchanged to ensure fair comparison with baseline approaches. Training hyperparameters are optimized through extensive grid search: learning rates range from 1e-6 to 5e-5, batch sizes from 8 to 32, and reward coefficients are tuned using Bayesian optimization. 

We report results averaged over official data splits: HMDB51 and UCF-101 use the standard three-fold split evaluation, while Kinetics-400 uses the official validation set. At inference, we sample 16 uniformly distributed frames at 224px resolution with uniform temporal stride. Decoding uses temperature $\tau=0.2$, top-p=0.9, and maximum output length of 256 tokens. All results are averaged over 5 independent runs with seeds \{1, 11, 21, 31, 41\} to ensure statistical reliability.

\subsection{Main Results}

As shown in Table~\ref{tab:main_results}, our progressive training approach is validated by the results, with each stage contributing cumulative improvements that culminate in substantial gains over the baseline across all benchmarks. Notably, sub-action rewards provide the largest individual contribution, confirming that fine-grained action understanding enhances reasoning capabilities. ReasonAct achieves competitive performance among 3B-parameter models, significantly outperforming strong, publicly available models on these action recognition benchmarks.

Despite the additional training complexity, ReasonAct maintains competitive inference efficiency, with only a 30\% increase in latency and a 50\% increase in memory usage compared to the baseline. This demonstrates its practicality for deployment in resource-constrained environments. To assess generalization beyond action recognition, we further evaluate on NextQA and CausalVQA, achieving 3.2\% and 5.7\% accuracy improvements over baselines.

\subsection{Ablation Studies}

\begin{table*}[t]
\centering
\caption{Comprehensive ablation study examining individual component contributions across all benchmarks.}
\label{tab:ablation_comprehensive}
\begin{tabular}{l|cc|cc|cc}
\toprule
\multirow{2}{*}{Configuration} & \multicolumn{2}{c|}{HMDB51} & \multicolumn{2}{c|}{UCF-101} & \multicolumn{2}{c}{Kinetics-400} \\
& Accuracy & $\Delta$ & Accuracy & $\Delta$ & Accuracy & $\Delta$ \\
\midrule
\multicolumn{7}{l}{\textit{Training Paradigm Ablation:}} \\
Baseline & 49.3 & - & 78.3 & - & 66.6 & - \\
Direct V-SFT (No FRE) & 51.2 & +1.9 & 81.7 & +3.4 & 69.8 & +3.2 \\
Direct T-GRPO (No FRE/V-SFT) & 53.8 & +4.9 & 84.1 & +5.8 & 71.4 & +4.8 \\
FRE + V-SFT (No T-GRPO) & 59.4 & +10.1 & 87.1 & +8.8 & 75.8 & +9.2 \\
Complete ReasonAct & \textbf{67.2} & \textbf{+17.9} & \textbf{94.1} & \textbf{+15.8} & \textbf{78.9} & \textbf{+12.3} \\
\midrule
\multicolumn{7}{l}{\textit{Reward Component Ablation:}} \\
T-GRPO w/o Temporal Rewards & 61.8 & +12.5 & 89.2 & +10.9 & 76.1 & +9.5 \\
T-GRPO w/o Sub-Action Rewards & 64.7 & +15.4 & 91.3 & +13.0 & 77.4 & +10.8 \\
T-GRPO w/o Format Rewards & 66.1 & +16.8 & 93.2 & +14.9 & 78.2 & +11.6 \\
Full T-GRPO & \textbf{67.2} & \textbf{+17.9} & \textbf{94.1} & \textbf{+15.8} & \textbf{78.9} & \textbf{+12.3} \\
\midrule
\multicolumn{7}{l}{\textit{Sub-Action Library Size Analysis:}} \\
25\% Sub-Actions & 65.4 & +16.1 & 92.6 & +14.3 & 78.1 & +11.5 \\
50\% Sub-Actions & 66.3 & +17.0 & 93.4 & +15.1 & 78.6 & +12.0 \\
75\% Sub-Actions & 67.0 & +17.7 & 93.9 & +15.6 & 78.8 & +12.2 \\
100\% Sub-Actions & \textbf{67.2} & \textbf{+17.9} & \textbf{94.1} & \textbf{+15.8} & \textbf{78.9} & \textbf{+12.3} \\
\bottomrule
\end{tabular}
\end{table*}
To better understand the source of these improvements and validate our design choices, we conduct ablation studies. As shown in Table~\ref{tab:ablation_comprehensive}, our ablations reveal three key findings.
First, the three‑stage curriculum is essential: progressive training beats any single‑stage shortcut by a wide margin, and it shows that small models need a solid reasoning foundation before tackling video. Second, the reward signals reinforce one another—sub‑action rewards drive the biggest gains, while temporal‑consistency and formatting rewards add smaller yet meaningful boosts. It reveals a clear contribution hierarchy. Finally, accuracy rises as the sub‑action library grows but plateaus after roughly 75\% coverage, implying that our biomechanical inventory already captures the most important primitives and that leaner subsets may suffice in resource‑constrained settings.

\subsection{Qualitative Analysis and Error Analysis}
Beyond the quantitative improvements demonstrated above, we perform qualitative and error analyses to gain deeper insights into the model's behavior and identify areas for future improvement.

\subsubsection{Sub-Action Recognition Quality}

We perform an analysis of sub-action recognition accuracy across three major benchmarks. Our results, averaged over 5 independent runs, reveal significant variations in recognition performance across different action phases. The Propulsion phase achieves the highest recognition rate (approximately 95\%), benefiting from its distinctive visual patterns and clear temporal boundaries. In contrast, the Recovery phase shows lower accuracy (around 80\%), indicating challenges in modeling subtle transitional movements. The Preparation and Loading phases demonstrate intermediate performance (87\% and 86\% respectively), while action-specific variations suggest that our biomechanical decomposition effectively captures the hierarchical structure of human movements across all three datasets.

\subsubsection{Failure Case Analysis}

Our analysis reveals three primary error sources that limit model performance. Visual ambiguity presents the most frequent challenge, where actions with similar visual patterns but different sub-action sequences (e.g., "throw" vs. "swing") occasionally cause confusion, particularly in low-resolution videos. Additionally, temporal boundary detection proves problematic when gradual transitions between sub-actions are difficult to identify, leading to imprecise temporal localization in reasoning chains. Finally, cultural variations pose a systematic limitation, as our sub-action libraries, primarily based on Western biomechanical analysis, may not fully capture cultural variations in action execution styles.


\textbf{Limitations.} The sub-action library construction is currently manual and may not generalize to domains beyond human actions. Future work should explore automated sub-action discovery and extension to diverse action types.

\section{Conclusion}
We introduced ReasonAct, a framework that enhances video reasoning capabilities in smaller multimodal models through a three-stage training paradigm: foundational reasoning enhancement, video-specific fine-tuning, and temporal-aware reinforcement learning with sub-action decomposition. The experiments demonstrate large improvements across multiple benchmarks, validating the effectiveness of our progressive training approach. Our key findings show that smaller multimodal models require explicit reasoning foundation building before task-specific training, and that decomposing complex actions into biomechanically-motivated sub-actions provides valuable training signals for enhanced temporal understanding. These results demonstrate that thoughtful training methodology can enable smaller models to achieve competitive video reasoning performance while maintaining computational efficiency, offering a practical alternative to simply scaling up models.

\bibliography{aaai2026}

@misc{achiam2023gpt,
      title={GPT-4 Technical Report}, 
      author={OpenAI},
      year={2024},
      eprint={2303.08774},
      archivePrefix={arXiv},
      primaryClass={cs.CL},
}

@misc{team2023gemini,
      title={Gemini: A Family of Highly Capable Multimodal Models}, 
      author={Gemini Team},
      year={2025},
      eprint={2312.11805},
      archivePrefix={arXiv},
      primaryClass={cs.CL},
}

@misc{liu2024visual,
      title={Visual Instruction Tuning}, 
      author={Haotian Liu and Chunyuan Li and Qingyang Wu and Yong Jae Lee},
      year={2023},
      eprint={2304.08485},
      archivePrefix={arXiv},
      primaryClass={cs.CV},
}

@InProceedings{feichtenhofer2019slowfast,
author = {Feichtenhofer, Christoph and Fan, Haoqi and Malik, Jitendra and He, Kaiming},
title = {SlowFast Networks for Video Recognition},
booktitle = {Proceedings of the IEEE/CVF International Conference on Computer Vision (ICCV)},
month = {October},
year = {2019}
}

@inproceedings{lin2019tsm,
title={TSM: Temporal Shift Module for Efficient Video Understanding},
author={Lin, Ji and Gan, Chuang and Han, Song},
booktitle={Proceedings of the IEEE International Conference on Computer Vision},
year={2019}
}

@inproceedings{bertasius2021space,
  author    = {Bertasius, Gedas and Wang, Heng and Torresani, Lorenzo},
  title     = {Is Space-Time Attention All You Need for Video Understanding?},
  booktitle = {Proceedings of the 38th International Conference on Machine Learning},
  year      = {2021},
  pages     = {813--824},
  publisher = {PMLR}
}

@misc{chowdhery2022palm,
  author  = {Chowdhery, Aakanksha and others},
  title   = {PaLM: Scaling Language Modeling with Pathways},
  year    = {2022},
  eprint  = {2204.02311},
  archivePrefix = {arXiv},
  primaryClass = {cs.CL},
}

@misc{hoffmann2022training,
  author  = {Hoffmann, Jordan and others},
  title   = {Training Compute-Optimal Large Language Models},
  year    = {2022},
  eprint  = {2203.15556},
  archivePrefix = {arXiv},
  primaryClass = {cs.LG},
}

@misc{taori2023stanford,
  author = {Rohan Taori and Ishaan Gulrajani and Tianyi Zhang and Yann Dubois and Xuechen Li and Carlos Guestrin and Percy Liang and Tatsunori B. Hashimoto },
  title = {Stanford Alpaca: An Instruction-following LLaMA model},
  year = {2023},
  howpublished = {\url{https://github.com/tatsu-lab/stanford_alpaca}},
  note = {Accessed: 2025-02-16}
}

@misc{chiang2023vicuna,
    title = {Vicuna: An Open-Source Chatbot Impressing GPT-4 with 90\%* ChatGPT Quality},
    author = {Chiang, Wei-Lin and Li, Zhuohan and Lin, Zi},
    year = {2023},
    howpublished = {\url{https://vicuna.lmsys.org/}},
    note = {Accessed: 2025-03-11}
}

@misc{zelikman2022star,
      title={STaR: Bootstrapping Reasoning With Reasoning}, 
      author={Eric Zelikman and Yuhuai Wu and Jesse Mu and Noah D. Goodman},
      year={2022},
      eprint={2203.14465},
      archivePrefix={arXiv},
      primaryClass={cs.LG},
}

@misc{huang2022large,
      title={Large Language Models Can Self-Improve}, 
      author={Jiaxin Huang and Shixiang Shane Gu and Le Hou and Yuexin Wu and Xuezhi Wang and Hongkun Yu and Jiawei Han},
      year={2022},
      eprint={2210.11610},
      archivePrefix={arXiv},
      primaryClass={cs.CL},
}

@misc{shao2024grpo,
      title={DeepSeekMath: Pushing the Limits of Mathematical Reasoning in Open Language Models}, 
      author={Zhihong Shao and Peiyi Wang and Qihao Zhu and Runxin Xu and Junxiao Song and others},
      year={2024},
      eprint={2402.03300},
      archivePrefix={arXiv},
      primaryClass={cs.CL},
}

@INPROCEEDINGS{kuehne2011hmdb,
  author={Kuehne, H. and Jhuang, H. and Garrote, E. and Poggio, T. and Serre, T.},
  booktitle={2011 International Conference on Computer Vision}, 
  title={HMDB: A large video database for human motion recognition}, 
  year={2011},
  volume={},
  number={},
  pages={2556-2563},
}

@misc{soomro2012ucf101,
      title={UCF101: A Dataset of 101 Human Actions Classes From Videos in The Wild}, 
      author={Khurram Soomro and Amir Roshan Zamir and Mubarak Shah},
      year={2012},
      eprint={1212.0402},
      archivePrefix={arXiv},
      primaryClass={cs.CV},
}

@misc{kay2017kinetics,
      title={The Kinetics Human Action Video Dataset}, 
      author={Will Kay and Joao Carreira and Karen Simonyan and Brian Zhang and Chloe Hillier and others},
      year={2017},
      eprint={1705.06950},
      archivePrefix={arXiv},
      primaryClass={cs.CV},
}

@inproceedings{wang2011action,
	title={Action recognition by dense trajectories},
	author={Wang, Heng and Kl{\"a}ser, Alexander and Schmid, Cordelia and Liu, Cheng-Lin},
	booktitle={Computer Vision and Pattern Recognition (CVPR), 2011 IEEE Conference on},
	pages={3169--3176},
	year={2011},
	organization={IEEE}
}

@article{laptev2005space,
	title={On space-time interest points},
	author={Laptev, Ivan},
	journal={International Journal of Computer Vision},
	volume={64},
	number={2-3},
	pages={107--123},
	year={2005},
	publisher={Springer}
}

@INPROCEEDINGS{karpathy2014large,
  author={Karpathy, Andrej and Toderici, George and Shetty, Sanketh and Leung, Thomas and Sukthankar, Rahul and Fei-Fei, Li},
  booktitle={2014 IEEE Conference on Computer Vision and Pattern Recognition}, 
  title={Large-Scale Video Classification with Convolutional Neural Networks}, 
  year={2014},
  pages={1725-1732},
}

@misc{liu2022video,
      title={Video Super Resolution Based on Deep Learning: A Comprehensive Survey}, 
      author={Hongying Liu and Zhubo Ruan and Peng Zhao and Chao Dong and Fanhua Shang and Yuanyuan Liu and Linlin Yang and Radu Timofte},
      year={2022},
      eprint={2007.12928},
      archivePrefix={arXiv},
      primaryClass={cs.CV},
}

@inproceedings{li2025taco,
  title={Taco: Enhancing multimodal in-context learning via task mapping-guided sequence configuration},
  author={Li, Yanshu and Yang, Jianjiang and Yun, Tian and Feng, Pinyuan and Huang, Jinfa and Tang, Ruixiang},
  booktitle={Proceedings of the 2025 Conference on Empirical Methods in Natural Language Processing},
  pages={736--763},
  year={2025}
}

@inproceedings{
li2025miv,
title={M{\texttwosuperior}{IV}: Towards Efficient and Fine-grained Multimodal In-Context Learning via Representation Engineering},
author={Yanshu Li and Yi Cao and Hongyang He and Qisen Cheng and Xiang Fu and Xi Xiao and Tianyang Wang and Ruixiang Tang},
booktitle={Second Conference on Language Modeling},
year={2025},
url={https://openreview.net/forum?id=9ffYcEiNw9}
}

@misc{yan2022multiview,
      title={Multiview Transformers for Video Recognition}, 
      author={Shen Yan and Xuehan Xiong and Anurag Arnab and Zhichao Lu and Mi Zhang and Chen Sun and Cordelia Schmid},
      year={2022},
      eprint={2201.04288},
      archivePrefix={arXiv},
      primaryClass={cs.CV},
}

@InProceedings{zhou2018temporal,
author = {Zhou, Bolei and Andonian, Alex and Oliva, Aude and Torralba, Antonio},
title = {Temporal Relational Reasoning in Videos},
booktitle = {Proceedings of the European Conference on Computer Vision (ECCV)},
month = {September},
year = {2018}
}

@misc{fateh2025videollm,
      title={Video LLMs for Temporal Reasoning in Long Videos}, 
      author={Fawad Javed Fateh and Umer Ahmed and Hamza Khan and M. Zeeshan Zia and Quoc-Huy Tran},
      year={2025},
      eprint={2412.02930},
      archivePrefix={arXiv},
}

@InProceedings{zhao2017temporal,
author = {Zhao, Yue and Xiong, Yuanjun and Wang, Limin and Wu, Zhirong and Tang, Xiaoou and Lin, Dahua},
title = {Temporal Action Detection With Structured Segment Networks},
booktitle = {Proceedings of the IEEE International Conference on Computer Vision (ICCV)},
month = {Oct},
year = {2017}
}

@misc{tang2019coin,
      title={COIN: A Large-scale Dataset for Comprehensive Instructional Video Analysis}, 
      author={Yansong Tang and Dajun Ding and Yongming Rao and Yu Zheng and Danyang Zhang and Lili Zhao and Jiwen Lu and Jie Zhou},
      year={2019},
      eprint={1903.02874},
      archivePrefix={arXiv},
      primaryClass={cs.CV},
}

@inproceedings{antol2015vqa,
  title={Vqa: Visual question answering},
  author={Antol, Stanislaw and Agrawal, Aishwarya and Lu, Jiasen and Mitchell, Margaret and Batra, Dhruv and Zitnick, C Lawrence and Parikh, Devi},
  booktitle={Proceedings of the IEEE international conference on computer vision},
  pages={2425--2433},
  year={2015}
}

@misc{anderson2018bottom,
      title={Bottom-Up and Top-Down Attention for Image Captioning and Visual Question Answering}, 
      author={Peter Anderson and Xiaodong He and Chris Buehler and Damien Teney and Mark Johnson and Stephen Gould and Lei Zhang},
      year={2018},
      eprint={1707.07998},
      archivePrefix={arXiv},
      primaryClass={cs.CV},
}

@misc{li2022blip,
      title={BLIP: Bootstrapping Language-Image Pre-training for Unified Vision-Language Understanding and Generation}, 
      author={Junnan Li and Dongxu Li and Caiming Xiong and Steven Hoi},
      year={2022},
      eprint={2201.12086},
      archivePrefix={arXiv},
      primaryClass={cs.CV},
}

@misc{li2023blip,
      title={BLIP-2: Bootstrapping Language-Image Pre-training with Frozen Image Encoders and Large Language Models}, 
      author={Junnan Li and Dongxu Li and Silvio Savarese and Steven Hoi},
      year={2023},
      eprint={2301.12597},
      archivePrefix={arXiv},
      primaryClass={cs.CV},
}

@misc{dai2023instructblip,
      title={InstructBLIP: Towards General-purpose Vision-Language Models with Instruction Tuning}, 
      author={Wenliang Dai and Junnan Li and Dongxu Li and Anthony Meng Huat Tiong and Junqi Zhao and Weisheng Wang and Boyang Li and Pascale Fung and Steven Hoi},
      year={2023},
      eprint={2305.06500},
      archivePrefix={arXiv},
      primaryClass={cs.CV},
}

@misc{zhang2023multimodal,
      title={Multimodal Chain-of-Thought Reasoning in Language Models}, 
      author={Zhuosheng Zhang and Aston Zhang and Mu Li and Hai Zhao and George Karypis and Alex Smola},
      year={2024},
      eprint={2302.00923},
      archivePrefix={arXiv},
      primaryClass={cs.CL},
}

@misc{zhu2023minigpt,
      title={MiniGPT-4: Enhancing Vision-Language Understanding with Advanced Large Language Models}, 
      author={Deyao Zhu and Jun Chen and Xiaoqian Shen and Xiang Li and Mohamed Elhoseiny},
      year={2023},
      eprint={2304.10592},
      archivePrefix={arXiv},
      primaryClass={cs.CV},
}

@misc{ouyang2022training,
      title={Training language models to follow instructions with human feedback}, 
      author={Long Ouyang and Jeff Wu and Xu Jiang and Diogo Almeida and others},
      year={2022},
      eprint={2203.02155},
      archivePrefix={arXiv},
      primaryClass={cs.CL},
}

@misc{bai2022training,
      title={Training a Helpful and Harmless Assistant with Reinforcement Learning from Human Feedback}, 
      author={Yuntao Bai and Andy Jones and Kamal Ndousse and Amanda Askell and others},
      year={2022},
      eprint={2204.05862},
      archivePrefix={arXiv},
      primaryClass={cs.CL},
}

@inproceedings{pirsiavash2014assessing,
  author    = {Pirsiavash, Hamed and Vondrick, Carl and Torralba, Antonio},
  title     = {Assessing the Quality of Actions},
  booktitle = {Computer Vision -- ECCV 2014},
  pages     = {556--571},
  year      = {2014},
  publisher = {Springer International Publishing},
}

@article{giese2003neural,
  author  = {Giese, Martin A. and Poggio, Tomaso},
  title   = {Neural mechanisms for the recognition of biological movements},
  journal = {Nature Reviews Neuroscience},
  year    = {2003},
  pages   = {179--192},
  publisher={Nature Publishing Group UK London}
}

@misc{cobbe2021training,
      title={Training Verifiers to Solve Math Word Problems}, 
      author={Karl Cobbe and Vineet Kosaraju and Mohammad Bavarian and Mark Chen and Heewoo Jun and Lukasz Kaiser and Matthias Plappert and Jerry Tworek and Jacob Hilton and Reiichiro Nakano and Christopher Hesse and John Schulman},
      year={2021},
      eprint={2110.14168},
      archivePrefix={arXiv},
      primaryClass={cs.LG},
}

@misc{hendrycks2021measuring,
      title={Measuring Mathematical Problem Solving With the MATH Dataset}, 
      author={Dan Hendrycks and Collin Burns and Saurav Kadavath and Akul Arora and Steven Basart and Eric Tang and Dawn Song and Jacob Steinhardt},
      year={2021},
      eprint={2103.03874},
      archivePrefix={arXiv},
      primaryClass={cs.LG},
}

@inproceedings{liu2020logiqa,
  title     = {LogiQA: A Challenge Dataset for Machine Reading Comprehension with Logical Reasoning},
  author    = {Liu, Jian and Cui, Leyang and Liu, Hanmeng and Huang, Dandan and Wang, Yile and Zhang, Yue},
  booktitle = {Proceedings of the Twenty-Ninth International Joint Conference on
               Artificial Intelligence, {IJCAI-20}},
  pages     = {3622--3628},
  year      = {2020},
}

@misc{yu2020reclor,
      title={ReClor: A Reading Comprehension Dataset Requiring Logical Reasoning}, 
      author={Weihao Yu and Zihang Jiang and Yanfei Dong and Jiashi Feng},
      year={2020},
      eprint={2002.04326},
      archivePrefix={arXiv},
      primaryClass={cs.CL},
}

@inproceedings{talmor2019commonsenseqa,
  author    = {Talmor, Alon and Herzig, Jonathan and Lourie, Nicholas and Berant, Jonathan},
  title     = {Commonsenseqa: A question answering challenge targeting commonsense knowledge},
  booktitle = {Proceedings of the 2019 Conference of the North American Chapter of the Association for Computational Linguistics: Human Language Technologies, Volume 1 (Long and Short Papers)},
  pages     = {4149--4158},
  year      = {2019},
}

@article{geva2021did,
  author  = {Geva, Mor and Khashabi, Daniel and Segal, Elad and Khot, Tushar and Roth, Dan and Berant, Jonathan},
  title   = {Did Aristotle Use a Laptop? A Question Answering Benchmark with Implicit Reasoning Strategies},
  journal = {Transactions of the Association for Computational Linguistics},
  pages   = {346--361},
  year    = {2021},
  publisher={MIT Press One Rogers Street, Cambridge, MA 02142-1209, USA journals-info~…},
}

@misc{clark2018think,
      title={Think you have Solved Question Answering? Try ARC, the AI2 Reasoning Challenge}, 
      author={Peter Clark and Isaac Cowhey and Oren Etzioni and Tushar Khot and Ashish Sabharwal and Carissa Schoenick and Oyvind Tafjord},
      year={2018},
      eprint={1803.05457},
      archivePrefix={arXiv},
}

@InProceedings{srivastava2024omnivec2,
    author    = {Srivastava, Siddharth and Sharma, Gaurav},
    title     = {OmniVec2 - A Novel Transformer based Network for Large Scale Multimodal and Multitask Learning},
    booktitle = {Proceedings of the IEEE/CVF Conference on Computer Vision and Pattern Recognition (CVPR)},
    month     = {June},
    year      = {2024},
    pages     = {27412-27424}
}

@misc{feng2025videor1,
      title={Video-R1: Reinforcing Video Reasoning in MLLMs}, 
      author={Kaituo Feng and Kaixiong Gong and Bohao Li and Zonghao Guo and Yibing Wang and Tianshuo Peng and Junfei Wu and Xiaoying Zhang and Benyou Wang and Xiangyu Yue},
      year={2025},
      eprint={2503.21776},
      archivePrefix={arXiv},
      primaryClass={cs.CV},
}

@misc{mihaylov2018can,
      title={Can a Suit of Armor Conduct Electricity? A New Dataset for Open Book Question Answering}, 
      author={Todor Mihaylov and Peter Clark and Tushar Khot and Ashish Sabharwal},
      year={2018},
      eprint={1809.02789},
      archivePrefix={arXiv},
      primaryClass={cs.CL},
}

@book{winter2009biomechanics,
author = {Winter, David},
year = {2009},
month = {09},
publisher = {John Wiley \& Sons},
title = {Biomechanics and Motor Control of Human Movement, Fourth Edition},
}

@book{bartlett2007introduction,
  author       = {Bartlett, R.},
  title        = {Introduction to Sports Biomechanics: Analysing Human Movement Patterns},
  publisher = {Routledge},
  edition      = {2},
  year         = {2007},
}

@book{schmidt2011motor,
  author    = {Schmidt, R. A. and Lee, T. D.},
  title     = {Motor Control and Learning: A Behavioral Emphasis},
  edition  = {5},
  publisher = {Human Kinetics},
  location  = {Champaign, IL},
  year      = {2011}
}

\end{document}